\documentclass[letterpaper]{article} 
\usepackage{aaai25}  
\usepackage{times}  
\usepackage{helvet}  
\usepackage{courier}  
\usepackage[hyphens]{url}  
\usepackage{graphicx} 
\usepackage{natbib}  
\usepackage{caption} 
\usepackage{algorithm}
\usepackage{algorithmic}
\usepackage{subfigure}
\usepackage{newfloat}
\usepackage{listings}
\DeclareCaptionStyle{ruled}{labelfont=normalfont,labelsep=colon,strut=off} 
\floatstyle{ruled}
\newfloat{listing}{tb}{lst}{}
\floatname{listing}{Listing}
\title{Expand VSR Benchmark for VLLM to Expertize in Spatial Rules}
\author{
    Peijin Xie\textsuperscript{\rm 1}, 
    Lin Sun \textsuperscript{\rm 2}, 
    Bingquan Liu \textsuperscript{\rm 1}\thanks{Corresponding author}, 
    Dexin Wang \textsuperscript{\rm 2}, 
    Xiangzheng Zhang \textsuperscript{\rm 2}, 
    Chengjie Sun \textsuperscript{\rm 1}, 
    Jiajia Zhang \textsuperscript{\rm 3}
}
\affiliations{

    Faculty of Computing, Harbin Institute of Technology, Harbin, 150001, China\textsuperscript{\rm 1},\\ 
    360 Search Department, Beijing, 100020, China \textsuperscript{\rm 2}, \\
    School of Computer Science and Technology, Harbin Institute of Technology, Shenzhen, 518055, China\textsuperscript{\rm 3},\\ 
    
    xpj@stu.hit.edu.cn, \{liubq,sunchengjie,zhangjiajia\}@hit.edu.cn, \\ \{sunlin1, wangdexin, zhangxiangzheng\}@360.cn
}

\usepackage{natbib} 
\begin{document}

\maketitle

\begin{abstract}
Distinguishing spatial relations is a basic part of human cognition which requires fine-grained perception on cross-instance.  
Although benchmarks like MME, MMBench and SEED  comprehensively have evaluated various capabilities which already include visual spatial reasoning(VSR).
There is still a lack of sufficient quantity and quality evaluation and optimization datasets for Vision Large Language Models(VLLMs) specifically targeting visual positional reasoning. 
To handle this, we first diagnosed current VLLMs with the VSR dataset and proposed a unified test set.
We found current VLLMs to exhibit a contradiction of over-sensitivity to language instructions and under-sensitivity to visual positional information.
By expanding the original benchmark from two aspects of tunning data and model structure, we mitigated this phenomenon. 
To our knowledge, we expanded spatially positioned image data controllably using diffusion models for the first time and integrated original visual encoding(CLIP) with other 3 powerful visual encoders(SigLIP, SAM and DINO).
After conducting combination experiments on scaling data and models, we obtained a VLLM VSR Expert(VSRE) that not only generalizes better to different instructions but also accurately distinguishes differences in visual positional information. 
VSRE achieved over a 27\% increase in accuracy on the VSR test set. 
It becomes a performant VLLM on the position reasoning of both the VSR dataset and relevant subsets of other evaluation benchmarks. 
We open-sourced the expanded model with data and Appendix at \url{https://github.com/peijin360/vsre} and hope it will accelerate advancements in VLLM on VSR learning.
\end{abstract}

%
\section{Introduction}

Reasoning on spatial relations is a basic part of human cognition which requires fine-grained perception on cross-instance. 
Traditional classification benchmark VSR ~\citep{vsr}  has proposed a controlled probing dataset testing vision language models' capabilities of discrimination on spatial relations with natural image-text pairs. 
But,as Large Language Models (LLMs) become research hotspots, traditional tasks have been incorporated into conversational QA scenarios. 
The new VSR task requires the model to not only accurately recognize visual positional information but also follow instructions to correctly answer questions.
Although models have evolved from traditional models with classification heads (ViLT, VisualBERT, and LXMERT) to VLLMs (LLaVA, BLIP2, Qwen-VL, and GPT-4), a unified evaluation method and effective optimization approach for VLLM toward VSR is still lacking.

VLLM is the advanced version of LLM, which is enhanced to process and interpret multi-modal data. They leverage a powerful LLM as their cognitive engine to handle various tasks. 
Equipped with visual tokens, LLMs can perceive rich visual information and perform complex reasoning like VSR. 
However, most VLLMs often fall into the trap of hallucination problems.  
During the model's hallucination evaluation, the spatial relationships are precisely the type of relational hallucinations that are more challenging to object and attribute hallucinations~\citep{hallu}. 
Therefore, filling the blank of the evaluation and optimization of VLLM on VSR is of great significance for addressing the issue of hallucinations on relation.

We first re-evaluated the VLLM from scratch and diagnosed issues of inconsistent performance, hypersensitiveness on text prompts, insensitivity on vision information and answer bias. 
We found the absence of a unified instruction test has resulted in significant variance in model performance and the instability of instruction-following hinders the evaluation and optimization of VLLM's VSR capabilities.
Therefore, we proposed a unified instruction test set by expanding the VSR test set through both manual and GPT4-generated templates.

Similarly, we expand the training set with the same template pool. 
Trained with more diverse QA formats, the model achieves fundamentally better generalization in responding to VSR questions regardless of the question style. 
Considering a VSR expert requires more of the ability to distinguish and recognize visual spatial information rather than just correctly following instructions, we controllably augmented and repainted the visual training image to specific spatial relation concepts (like ``in'', ``on'' and ``under'') by diffusion model. 
The increased amount and diversity of image data strengthen the model's comprehension of visual spatial details. 
Furthermore, we expand the vision encoder CLIP~\citep{clip} with other hot encoders (SAM~\citep{sam}, DINOv2~\citep{dinov2} and SigLIP~\citep{siglip}) to obtain more vision perception ability that promotes model sensitivity to spatial information. 
After the experiments on scaling laws by data augmentation and ablation study on encoder combination, we obtained a VLLM VSR Expert(VSRE) that not only generalizes better to different text instructions but also accurately distinguishes differences in visual positional information.
Our expansion achieved over  27\% increase in accuracy on the VSR bench. 
The VSRE achieved the best performance on the position reasoning of both the VSR dataset and subsets of other evaluation benchmarks. 
Additionally, the phenomenon of answer bias has also been alleviated as the model's attention shifts towards the visual aspects.

The contribution of this work can be summarized as follows:

\begin{enumerate}
    \item We proposed the first unified VSR instruction test set and identified critical issues of inconsistent performance, hypersensitiveness on text, insensitivity on vision and answer bias.
    \item We expanded the training data  nearly 100 times by rewriting the text and repainting the image.
    \item We expanded the single visual backbone to a merged visual encoder which enhanced the VLLM's vision perception ability.
    \item  Our proposed VSRE overcomes the issues. It not only generalizes better to different text instructions but also accurately distinguishes differences in visual positional information.
\end{enumerate}
\begin{figure*}
\centering
\includegraphics[scale = 0.5]{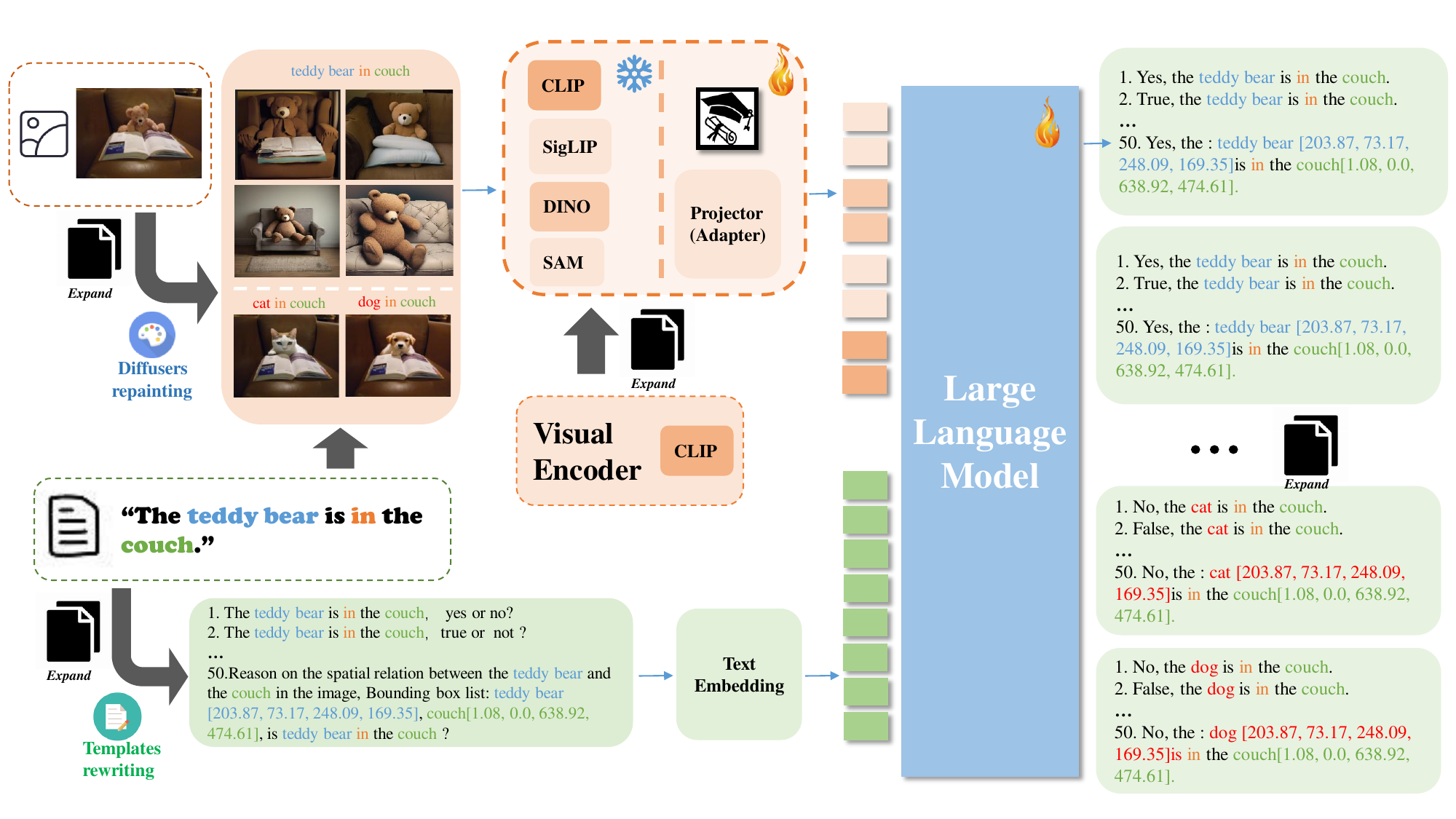}
  \caption{The overall expansion method through the training process. On the text branch, questions and answers are rewritten by temples in green blocks. On the image branch, the image inputs are repainted by the diffusion model through \textit{image to image}, \textit{text to image} and \textit{image inpainting} 3  methods in respectively 3 rows in the left orange block. And the expansion on vision encoder to a powerful merged one is shown in the middle orange block with dashed box. }
  \label{fig:overall}
\end{figure*}

\section{Related Works}
\subsection{VSR Tasks and Datasets}
The VSR dataset~\citep{vsr} is a controlled dataset that explicitly evaluates traditional pretrained vision language models like VisualBert, LXMERT and Vilt. 
It contains more than 10k natural text-image pairs with 66 types of spatial relations which is still insufficient for instruction tunning for VLLM. 
During the image data selection from MSCOCO~\citep{mscoco}, they used a contrastive caption generation approach and manual second-round validation to guarantee clear positional relationship features and data balance. 
As each image-text pair sample contains a triplet of [\textit{subject}(with bbox), \textit{relation}, \textit{object}(with bbox)], the VSR dataset has excellent extensibility under the QA scenario where VLLM fits.

Existing hot benchmarks like MME~\citep{mme}, MMbench~\citep{mmbench}, SEED~\citep{seed} test VLLMs' various capabilities and also have integrated assessments of VSR capability. 
However, the amount of their relevant subsets, only makes them suitable for evaluating model capabilities but insufficient for tuning the VLLM.

Previous works such as InstructBlip~\citep{instructblip}, LLaVA~\citep{improved,llava1.5,llava} and Mm1~\citep{mm1} transforms existing VQA benchmarks~\citep{vg, vqa} into instruction tuning data, showing marked VLLM performance improvements.
With this inspiration, we take well-controlled and balanced original VSR triplets as seeds, expand them dozens of times to vision instruction data to fill in the gaps between traditional visual language models and VLLM.

Several VLLMs have posted their performance on VSR. Due to the lack of a unified test template, their result seemed inconsistent through reproduction. 
We specifically discuss this issue in the Section \textbf{Re-evaluate from Scratch} \ref{re-eval}.

\subsection{Merged Vision Encoder}

Current works explored the benefits of multi-visual joint encoding across various perception and cognition tasks. 
Their ablation experiments identified the best combinations of popular vision features for specific tasks. 

Mixture of Features (MoF) ~\citep{eyes} demonstrated that integrating vision self-supervised learning DINOv2~\citep{dinov2} features with VLLMs can significantly enhance their visual grounding capabilities.  
MG-LLaVA~\citep{mgllava} enhances the model’s visual processing capabilities by incorporating a multi-granularity vision flow, which includes low-resolution, high-resolution, and object-centric features. 
Prism~\citep{prismatic} perform a head-to-head comparison between CLIP~\citep{clip}, SigLIP~\citep{siglip}, DINOv2, and a standard Vision Transformer pretrained for classification (on ImageNet-21K, finetuned on ImageNet-1K) and find that the backbones trained with vision-language contrastive objectives (i.e. CLIP and SigLIP) are significantly more performant than alternatives. 
Cambrian-1~\citep{cambrian} also explored 20 vision encoders and their combinations. 
They conclude that High-res encoders greatly enhance performance on chart\&vision-centric benchmarks. Combining multiple vision encoders, including  Self Supervise Learning models, enhances VLLM performance across various benchmarks, particularly in vision-centric tasks.

Motivated by their findings, we summarized their high-performing backbones(including SigLIP, DINOv2 and SAM~\citep{sam}) and followed their combination choice to expand our vision feature.  
After all, when it comes to intuitively understanding VSR tasks, precise visual detail perception and discrimination capabilities seemed more essential than complex semantic reasoning abilities. 
This necessitates that the model's visual feature processing component be more sensitive to visual positional information and possess more specialized spatial encoding capabilities.

\subsection{Controled Image Generation}
Diffusion models for controlled image generation offer a powerful and flexible approach to creating images that meet specific criteria. 
By incorporating conditioning information into the denoising process, models can generate highly controlled and precise outputs. 
They represent a powerful approach for image repainting, leveraging the systematic denoising process to generate high-quality, contextually consistent images. 
In this work, we utilized the most 3 popular applications to expand the image data as follows:

\textbf{Text to Image} generates an image from a text description. 
The denoising process is guided by the text, and once the denoising process ends after a predetermined time steps, the image representation is decoded into an image.

\textbf{Image-Text to Image} is similar to text-to-image, but in addition to a prompt, an initial image is encoded to latent space then the noise is added to it. Then the model takes a prompt and the noisy latent image, predicts the noise, and removes the predicted noise from the initial latent image.

\textbf{Image Inpainting} replaces or edits specific areas of an image. This makes it a useful tool to replace an image area with something entirely new. 
It relies on a mask to determine which regions of an image to fill in; the area to inpaint is represented by white pixels and the area to keep is represented by black. The white pixels are filled in by the prompt.


\section{Re-evaluation from Scratch}
\label{re-eval}
In this section, we re-evaluate the performance of popular VLLMs on the VSR test from scratch and identify the problem of inconsistent performance, hypersensitiveness on text, insensitivity on vision and answer bias that hinder the evaluation. Due to page limitations, we provide a brief overview of our re-evaluation process as follows, more details are shown in Appendix Material Section 1.

We first provide the most comprehensive summary 
including over 120 attempts across more than 30 models. 
During the summary constitution, We identified inconsistencies in model performance, such as the model LLaVA1.5 having nearly a 20\% lower accuracy in MiniGPTv2's reproduction(51\%)~\citep{minigptv2}  compared to the result in Prism(71\%)~\citep{prismatic}. 

Then we selected the most commonly used VLLM architecture, LLaVA1.5, for prompt engineering experiments with 69 templates. 
We found that hypersensitiveness on text prompts caused the inconsistency in performance. 
When asked about positional information, the model's accuracy is significantly affected by factors such as the questioning format, the insertion or deletion of specific phrases, and the order of certain words.
Similarly, through case studies, we found that insensitive visual features struggle to distinguish positional categories.

Furthermore, the model exhibited severe response biases for most template answers. 
Questions with ``yes'' answers have a significantly higher accuracy rate compared to those with ``no''.
We suspect that may be due to the co-occurrence of subject and object concepts in both question text and images.
This co-occurrence phenomenon may confuse the model, leading it to hastily provide ``yes'' answers based on co-occurrence rather than focusing on the actual visual spatial relationship.

To conclude, the over-sensitivity to language instructions caused large variance and inconsistency in model performance, while under-sensitivity to visual information led to insufficient perception of spatial relations, affecting answer accuracy and introducing response bias.






\begin{figure}
\centering
\includegraphics[scale = 0.25]{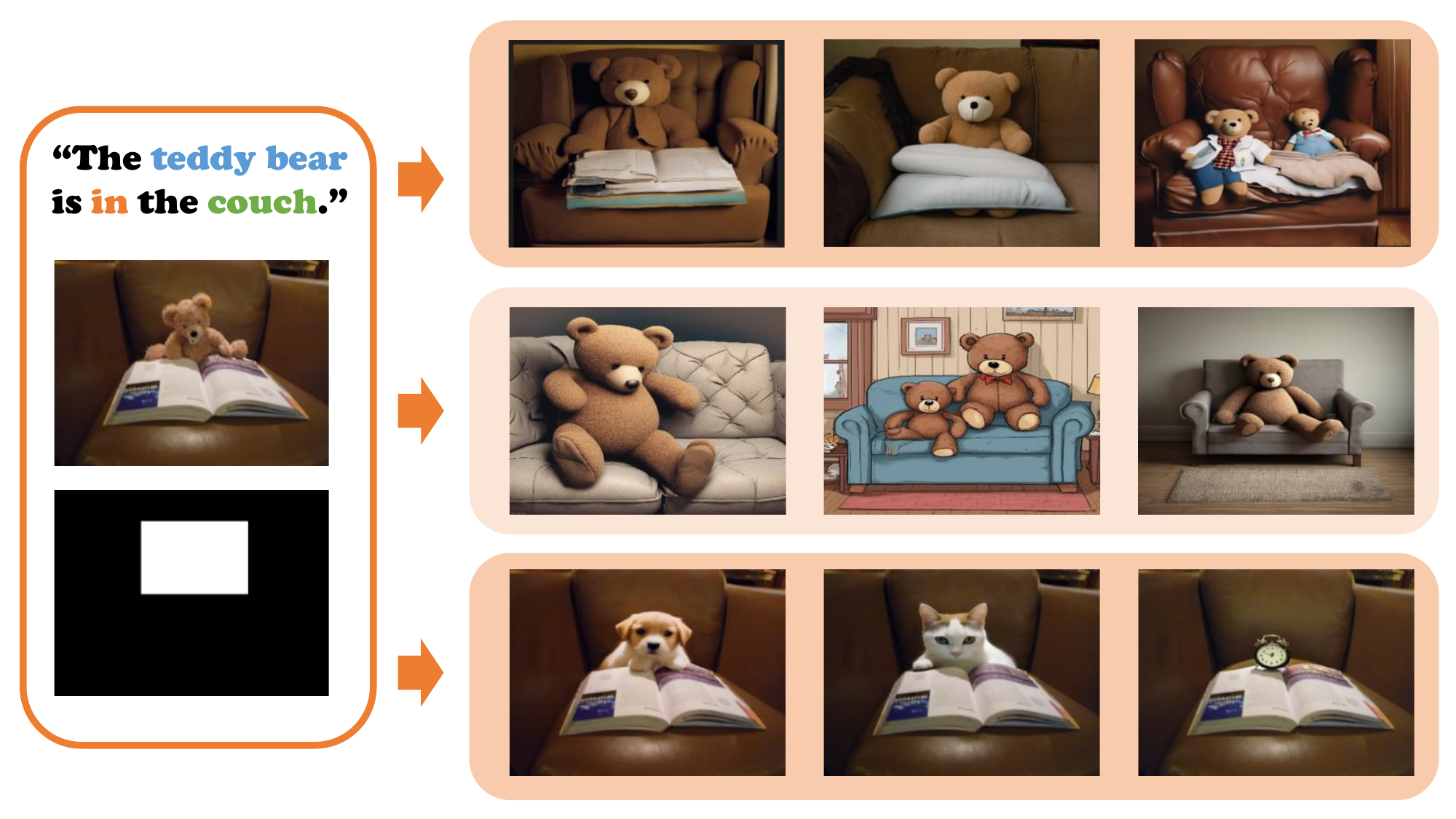}
  \caption{Examples of 3 settings of image-to-image(first row), text-to-image(middle row), and inpainting(last row) through the repainting process with the original image-text pair and mask inputs on the left.}
  \label{fig:diffusion}
\end{figure}

\section{Expansion for Spatial Expert}
To overcome the issues and obtain a Spatial Expert, we arrange our expansion method through VLLM's pertaining and instruction tunning pipeline in Figure \ref{fig:overall}.

  

\subsection{Expansion on Text Data }
Inspired by the success of visual instruction tuning, we prepare the test training data from the original VSR training [\textit{subject}, \textit{relation}, \textit{object}] triplets extracted by spaCy\footnote{\url{https://spacy.io/}}. 
We hope that after instruction fine-tuning(IFT) with augmented text data, the model, having been exposed to more diverse forms of Q\&A, can reduce its sensitivity to question text prompts and exhibit higher generalization on instruction following when faced with various questioning formats.
We expand the text QA data through both manual and GPT-4o generated prompt templates. 
To be exact, we select the top 30 templates manually in the re-evaluation process and append 20 GPT4-o generated ones.

\subsection{Expansion on Image Data }
To enhance the perception of position on VLLM, we provide them with dozens of times more image inputs with spatial relation concepts than ever.
Motivated by the success on LLaVA pretrained on GPT4 generated captions, inversely, we freeze the text captions as prompts and utilize image generation diffuser SDXL~\citep{sdxl} to generate images. 
More details and samples of the generation are posted in the Appendix Material Section 2.

The general image generation under prompt control ensures the display of specific positional relationship information between two entities, such as ``The teddy bear is in the couch''.
We believe that an efficient principle for data augmentation is to generate images with as much diversity as possible while ensuring the original positional relationship semantics are maintained.
Therefore, we employed the most three general settings to sequentially increase diversity.

As shown in Figure \ref{fig:diffusion}, \textbf{Image-to-Image} repaint the picture slightly in the first row which also maintained the consistency of the color on the background and the white region in front of the bear. 
The differences might not be apparent after visual encoding.

Therefore, we armed our augmentation with \textbf{Text-to-Image} and \textbf{Image-Inpainting} strategic.
Text-to-Image generated a more diverse set of images including variations in background and couch color and image style in the second row of Figure\ref{fig:diffusion}. 
And in the last row, we utilized the bounding box as mask to inpaint the photo to make alter on \textit{subject} and \textit{object}. 
We replaced relatively small items with similar concepts like the bear with a dog, a cat and a clock. 
The revision helps prevent the model from learning biases on the distribution of \textit{subject} and \textit{object} combinations in the text.




\subsection{Expansion on Vision encoder}
\begin{figure}
\centering
\includegraphics[scale = 0.25]{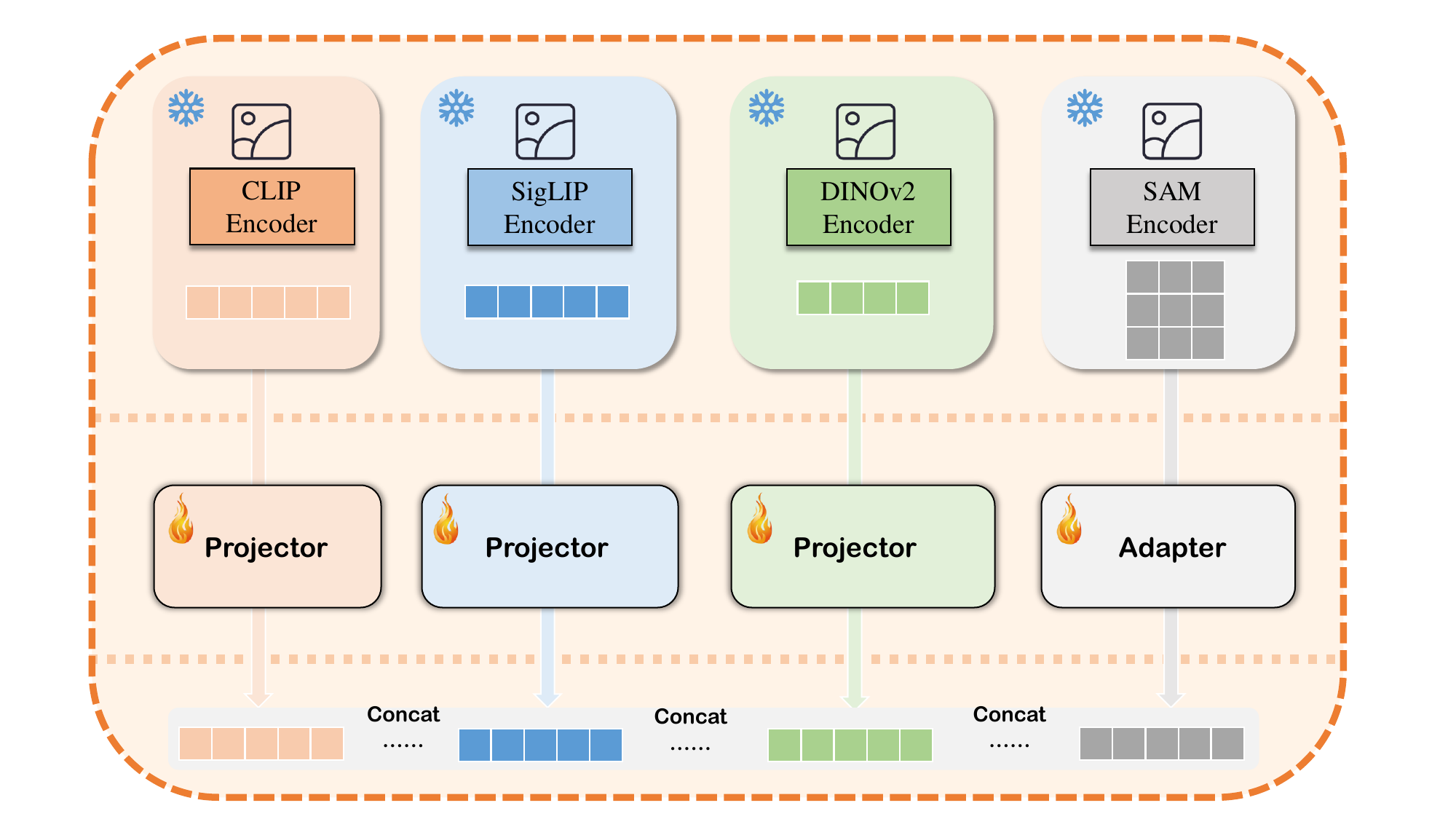}
  \caption{Illustration of the Merged Vision Encoder that concatenate multiple visual features aligned by projector or adapter respectively.}
  \label{fig:merged_encoder}
\end{figure}
With sufficient VSR text and image data support, we incorporated the most used visual backbone CLIP\footnote{\url{https://huggingface.co/openai/clip-vit-large-patch14-336}} with SigLIP\footnote{\url{https://huggingface.co/google/siglip-so400m-patch14-384}}, DINOv2\footnote{\url{https://huggingface.co/facebook/dinov2-base}} and SAM\footnote{\url{https://huggingface.co/facebook/sam-vit-base}} to fully explore the potential of combining visual features. 
The language-guided contrastive model CLIP and SigLIP benefit from the massive scale of noisy web image-text data. 
But self-supervised encoders DINOv2 and segmentation SAM may detect more fine-grained visual details which may profit the spatial reasoning process.
We then merge the pretrained backbones on the shelf to obtain a superior vision encoder focusing on visual spatial relations. 
Specifically, as shown in Figure\ref{fig:merged_encoder}, we design projectors and adapter to align  visual tokens,
concatenate these tokens along the feature dimension following  MOF, MG-LLaVA and Cambrian. 
Additional designs of the projector and adapter are shown in Appendix Section 3.

\section{Experiment}
In this section, we illustrate the details of the dataset creation and trainng\&inference arrangement. 
As for the baseline model, we adopt the most extensively adopted VLLM architecture LLaVA1.5 to verify our expansion methods. 
Similar to its training process, we also arrange our optimized experiment into 2 stages: 
(1) Pertaining stage to enhance the perception ability of visual spatial relation and understanding of positional text concepts. 
(2) Instruction fine-tune (IFT) stage to enhance instruction following ability then reduce the sensitivity to question text prompts and obtain higher generalization towards variant question formats.
\subsection{Datasets}

\subsubsection{Testing Datasets} 
To maintain consistency with previous evaluations, we used the VSR zero-shot test sets as the basis. 
We organized our test data into two sets:

(1)\textit{Test-G} random sampled a prompt from the 50 templates pool for each triplet to evaluate instruction-following generalization ability during spatial reasoning.

(2)\textit{Test-S} froze the template to the specific one ([\textit{caption}], True or false.) which performs the best and is the simplest during the re-evaluation process. 
The split aims to test the model's VSR capability on visual under its most proficient question-asking format.

\subsubsection{Training Datasets}
Excluding the test set, we collected more than 10k triplets with images from the original VSR dataset as a seed. Then we expand it several dozens of times into pre-train and IFT data as follow:(more details in Appendix Section 4)
\begin{itemize}
    \item Pre-training data: Under the three settings with a ratio of 5:3:2, we repainted the original images, expanding the quantity 20 to 100 times the original amount. We label the set as ``pre-100k'' for 100k pre-training data and ``pre-500k'' for 500k. 
    \item IFT data: We used general 50 prompt templates (30 manual and 20 GPT4-generated) to expand the 11k triplet data nearly 50 times to 500k, then name it as ``turn-g 500k''. Note that ``turn-s 11k” and ``turn-g 11k'' are unexpanded IFT data for comparison.  ``turn-g 11k''  used a randomly selected temple for each triplet and ``turn-s 11k'' used the specific template same as \textit{Test-S}.
\end{itemize}

\subsection{Training and  Inference}
To ensure fairness, we seed everything before the training and inference process. 
Throughout the entire training process, we froze all visual backbones. 
During the pre-training stage, we froze the LLM and only trained the adapter layers to encourage the model to learn better visual encoding of spatial details. 
In the fine-tuning stage, we unfroze the LLM, allowing it to participate in the training to enhance the model's instruction-following capability.
Finally, during the inference stage, we limited the length of the model's responses to only one new word. 
This word was used as the answer to binary questions; for example, words like ``True'', ``Yes'' or ``A'' were considered positive predictions, while ``False'', ``No'' or ``B'' were considered negative. 
Any other responses were directly judged as incorrect.

\section{Result and  analysis}
\begin{table}[]
    \footnotesize
    \centering
    
\begin{tabular}{cc|cc}
\hline
\begin{tabular}[c]{@{}c@{}}Pretrain\\ (Adapter)\end{tabular} & \begin{tabular}[c]{@{}c@{}}IFT\\ (Adapter + LLM)\end{tabular} & \begin{tabular}[c]{@{}c@{}}acc 7B\\ \textit{Test-G}/\textit{Test-S}\end{tabular} & \begin{tabular}[c]{@{}c@{}}acc 13B\\ \textit{Test-G}/\textit{Test-S}\end{tabular} \\ \hline
-              & -                     & 54.3 / 65.3        & 57.7 / 68.4           \\ \hline
-              & turn-g 11k             & -                 & -                 \\ \hline
turn-g 11k      & -                     & 57.3 / 65.7     & 59.2 / 68.2                 \\ \hline
-              & turn-s 11k             & -                 & -               \\ \hline
turn-s 11k      & -                     & 55.1 / 67.9     & 56.7 / 70.1                  \\ \hline
-              & turn-g 500k            & 58.3 / 69.5          & 62.5 /  71.4                  \\ \hline
pre-100k       & turn-g 500k            & 61.7 / 71.0             & 63.2 /  73.7                 \\ \hline
pre-200k       & turn-g 500k            & 64.1 / 73.3             & 65.8 /  74.9                 \\ \hline
pre-300k       & turn-g 500k            & 65.6 / 74.1             & 69.7 /  75.5                  \\ \hline
pre-400k       & turn-g 500k            & 66.7 / 73.5             & 70.3 /  75.7                  \\ \hline
pre-500k       & turn-g 500k            & 66.2 / 73.6             & 70.2 /  75.6                  \\ \hline
\begin{tabular}[c]{@{}c@{}}pre-400$k_1$\\ turn-s 11$k_3$\end{tabular}  & turn-g 500$k_2$    & 66.4 / 74.7  & 70.1 / 76.6                    \\ \hline

\end{tabular}
    \caption{Result of LLaVA1.5 7B and 13B on scaling training data experiment. We post the Test-G and Test-S accuracy (split through ``/'') by pretrained the adapter with data of the first column and instruct fine turning(IFT) both the adapter and LLM with the second column data.}
    \label{table:llava}
\end{table}

\begin{table*}[]
    \footnotesize
    \centering
\begin{tabular}{c|ccccccc}
\hline
LLM / VLLM                          & zreo-shot &turn-g 500k & +pre-100k & +pre-200k & +pre-300k & +pre-400k & +pre-500k \\ \hline
vicuna 7B                         & -         &56.4 / 64.0  &58.3 / 67.2          &60.2 / 68.2          &61.6 / 70.1          &63.1 / 72.6          &63.4 / 72.9          \\
vicuna 13B                        & -         &59.8 / 67.3  &62.4 / 69.1          &64.7 / 70.8          &65.7 / 73.5          &68.7 / 74.2          &69.2 / 74.2          \\
LLAMA2 7B                         & -         &57.1 / 61.7  &57.8 / 64.5          &61.3 / 67.3          &62.1 / 68.2          &63.3 / 69.9          &62.4 / 70.1          \\
LLAMA2 13B                        & -         &60.9 / 63.2  &61.8 / 67.8          &65.5 / 70.6          &67.0 / 72.4          &69.2 / 73.4          &68.8 / 74.4          \\
LLAMA3 8B                         & -         &61.5 / 65.4  &62.6 / 68.4          &65.8 / 70.5          &68.5 / 72.9          &70.0 / 74.1          &70.0 / 74.3          \\ \hline
Qwen-VL 7B                        &57.8 / 62.7     &63.4 / 68.2     &65.8 / 69.2    &66.1 / 71.7   &67.0 / 72.2    &67.9 / 73.6    &68.2 / 73.6          \\

\begin{tabular}[c]{@{}c@{}}BILP2 \\ (FlanT5XXL)\end{tabular}       &59.3 / 66.5    &64.2 / 69.6      &66.3 / 69.7    &68.0 / 71.1    &69.3 / 72.5    &70.4 / 73.9    &70.2 / 74.2          \\
\begin{tabular}[c]{@{}c@{}}InstructBLIP\\ (FlanT5XXL)\end{tabular} &54.4 / 63.1     &59.0 / 66.2      &62.5 / 67.8    &64.3 / 69.7    &66.9 / 71.8    &67.3 / 71.9    &68.2 / 72.3          \\ \hline
\end{tabular}
    \caption{Result of scaling data across other hot-spot LLM and VLLM on Test-G and Test-S (split through ``/''). The column names represent the data used for training, in sequence of first 3 as: no data for ``zero-shot'', only 500k turn-g data for ``turn-g 500k'', turn-g 500k plus pre-100k for ``+pre-100k''. 
    For LLMs, we randomly initialized the adapter weights and sequentially used the corresponding data to perform the pretrain and fine-tune processes.
    And for VLLMs, we combined the tuning and pre-train data sequentially for IFT.}
    \label{table:llms&vllm}
\end{table*}

\subsection{Scaling on Data}
We validated the effectiveness of our expanded data on LLaVA, further replaced the LLM and VLLM, the results showed that our data expansion adapted to various models.

Table \ref{table:llava} shows the result of LLaVA1.5 7B and 13B on scaling training data experiment. 
We pre-trained and tuned the existing projector(adapter) and LLM with their weights on-the-shelf using the augmented data. 
We found that during pre-training, adjusting the projector(adapter) with even just 11k of IFT data can lead to an increase in accuracy, but model failed to respond after tuning LLM by tiny 11k data. 
Therefore, we added the augmented 500k tune-g data and gradually increased the pre-train samples.
First, after training with the turn-g 500k IFT samples, the 7B model improved by 4.0\% on the \textit{Test-G} set, and 13B model improved by 4.8\%. 
Both versions achieved certain improvements on the \textit{Test-S}.
As the amount of pre-training data increased, the models continued to achieve better results. 
Although the accuracy gains became less noticeable when the data size increased beyond 400k, the 13B model ultimately achieved scores of 70.3 and 75.7 on each test.
This indicates that data expansion not only helps to enhance the model's generalization ability to answer various text questions but also improves the model's ability to discern positional information. 
Notably, in the last row, we added tuning data turns-s 11k at the end, meaning we first pre-trained with pre-400k, then tuned with turn-g 500k, and finally pre-trained again with turn-s (the training sequence is marked with subscripts number in the table). 
As a result, we found that the model's performance on the \textit{Test-S} further improved to 76.6.
This demonstrates that supplementing with relevant data has great potential for optimizing model performance on VSR.


Table \ref{table:llms&vllm} shows the result of scaling data across other hot-spot LLM and VLLM. 
We replaced the tuned LLM in LLaVA with other LLMs and randomly initialized the adapter for retraining.
Also, we combined the pretrain and the tune set to fine-tune other VLLMs together.
The results indicate that although the performance was not as good as continuing training on the trained LLaVA, our data augmentation method significantly improved other LLMs and VLLMs on VSR task. 
The accuracy of the model's responses improved consistently in both randomly asked general questions and fixed-format questions.

\subsection{ Scaling on Model}
\begin{figure}
\centering
	\begin{minipage}[t]{0.4\linewidth}
		\centering
		\includegraphics[scale = 0.24]{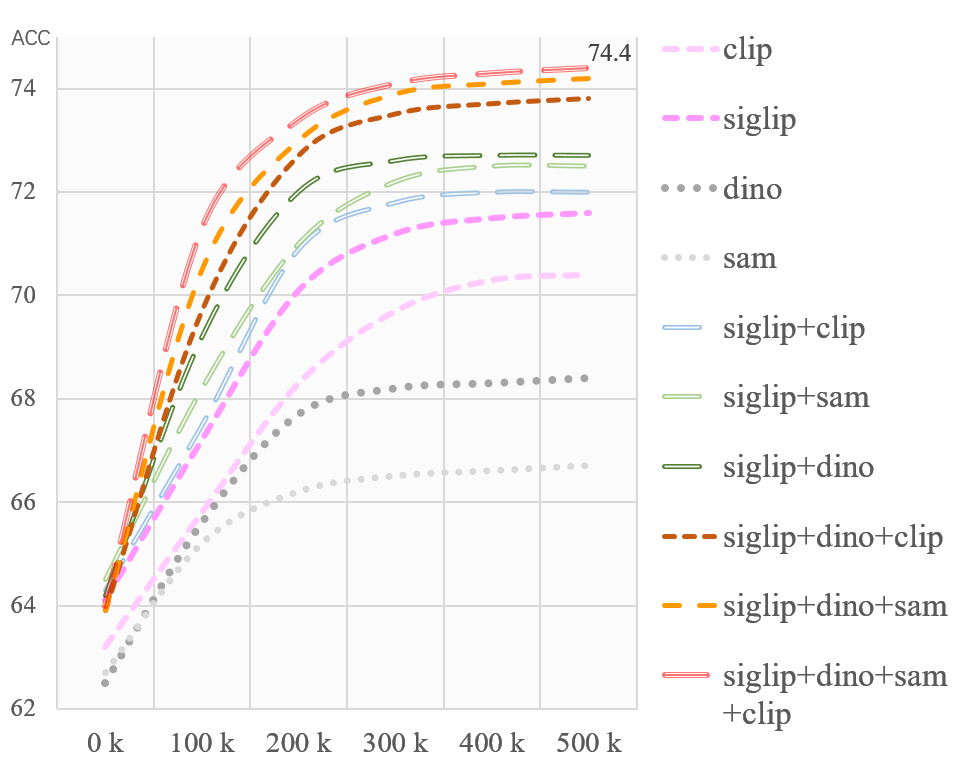}
	\end{minipage}
        \hspace{0.8cm}
	\begin{minipage}[t]{0.4\linewidth}
		\centering
		\includegraphics[scale = 0.24]{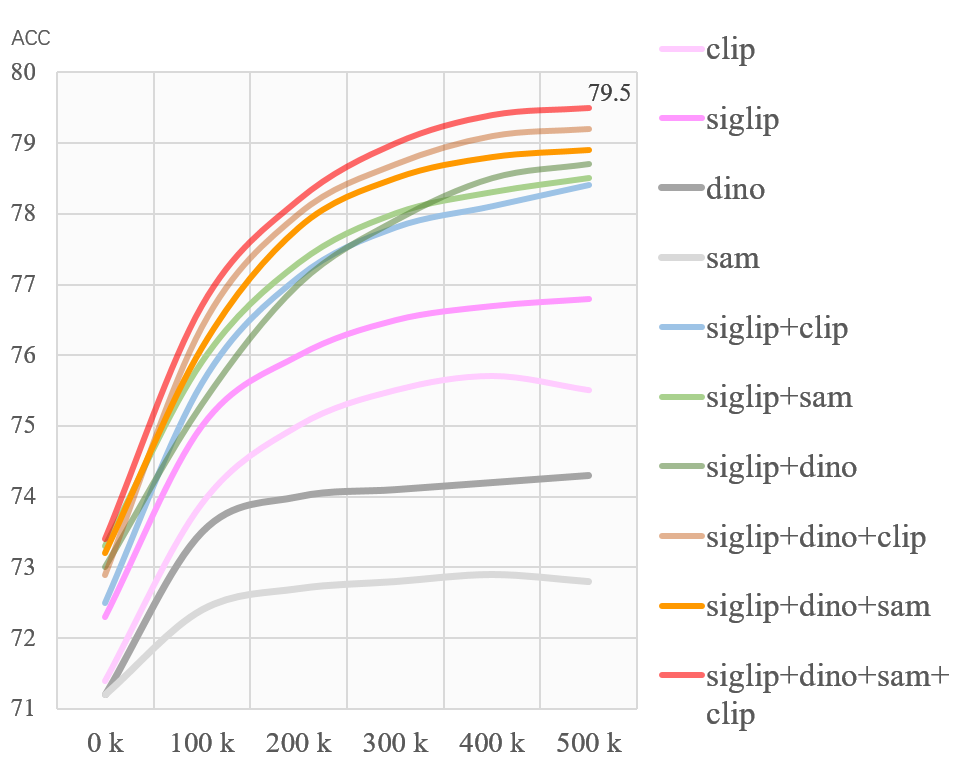}
	\end{minipage}
	\caption{Result of scaling vision model. We post the accuracy of \textit{Test-G} on the left in dashed lines and \textit{Test-S} on the right in solid lines.}
	\label{fig:rainbow}
\end{figure}
The 2 ``rainbows'' in Figure \ref{fig:rainbow} shows the result of experiments on scaling vision model.
The left ``rainbow'' illustrates the performance on  \textit{Test-G} set during the accumulation of pre-train data. 
In the comparison of the four individual visual backbones, SigLIP performed the best.
Then CLIP performed the next, followed by DINOv2 and SAM.
This reflects the superiority of language-supervised visual backbones in such multimodal tasks.
Then we freeze SigLIP and append other visual features. 
In the comparison of binary visual backbone combinations, SigLIP+DINO emerged as the fastest and best learner in deep green dashed line.
This indicates that the self-supervised backbone DINOv2 helps provide more detailed information, which aids the model in focusing on finer visual details in VSR QA scenarios.
We freeze the SigLIP+DINO and add CLIP and SAM respectively. 
In the comparison of triple backbones, CLIP brought fewer benefits compared to SAM. 
This may be due to the high feature overlap between CLIP and SigLIP, as they are both contrastive learning models supervised by language.
Finally, with the combined efforts of the four backbones, the accuracy of \textit{Test-G} reached the highest value of 74.4.

Similarly, the right ``rainbow'' shows the accuracy on the \textit{Test-S}. 
The combination results of each backbone are generally consistent with those of \textit{Test-G}, except that in the comparison of triple backbones, the appended CLIP outperforms SAM.
Another point is that compared to \textit{Test-G}, the model learns faster under the single backbone setup. 
This is likely due to the reduced requirements on the language question side with the use of a fixed template, which lowers the overall task difficulty. 
Consequently, a smaller amount of data is sufficient for the model to achieve learning saturation.
Nevertheless,  the scaling model method ultimately improved the accuracy of \textit{Test-S} to 79.
Moreover, it can be observed that a common characteristic of both test sets is that while adding new visual features results in accuracy gains, the incremental improvement in accuracy diminishes as the number of added visual features increases.
Finally, we designated the best-performing model with 4 backbones trained by the total data obtained as a Visual Spatial Reasoning Expert(VSRE).






\subsection{Result on Other Benchmarks} 
To verify the generalization of our expansion method, we tested our VSRE on the 
MME\footnote{\url{https://github.com/BradyFU/Awesome-Multimodal-Large-Language-Models}}, 
MMBench\footnote{\url{https://mmbench.opencompass.org.cn/home}} and 
SEEDv2\footnote{\url{https://github.com/AILab-CVC/SEED-Bench}}.
Table \ref{result:other3} show the comparison  of VSRE on the related subsets of other benchmarks.
Firstly, on the MME dataset, which also involves answering binary questions as VSR, VSRE achieved the highest score of 155.33, an improvement of more than 22 points compared to the optimized baseline LLaVA1.5 13B.
This indicates that our expanded data and model are not overfitted to the single VSR dataset but show robust performance across benchmarks.
Moreover, even though MMBench and SEEDv2 ask multiple-choice questions, VSRE still achieved the best results. 
Compared to the baseline LLaVA before optimization, the scores improved by 7.2 and 8.1 respectively.
This may be due to our diverse template-designed IFT dataset, which included question formats similar to multiple-choice questions.
Overall, the best performance among the dataset subsets demonstrates that our expansion method for VSR is not overfitting to similar data. 
Instead, it has genuinely learned visual position reasoning capabilities, handled more diverse text question formats and discerned visual positional information.


\begin{table}[]
\begin{tabular}{c|ccc}
\hline
Model         & MME        & MMBench & SEEDv2 \\ \hline
MiniGPT-4v2   & 43.33      & -        & 32.6 \\
Qwen-VL(chat) & 128.33     &47.2      & 40.3 \\
LLaVA1.5 13B  & 133.33     &57.6      & 38.5 \\
BLIP2         & 73.33      &58.4      & 36.2 \\  \hline
VSRE          & 155.00     &64.8      & 46.6    \\ \hline
\end{tabular}

    \caption{The comparison results of VSRE on the related subsets of other datasets including MME, MMBench and SEEDv2.}
    \label{result:other3}
\end{table}

\subsection{More Sensitive Vision Features} 
\begin{figure}

	\begin{minipage}[t]{0.4\linewidth}
		\centering
		\includegraphics[scale = 0.26]{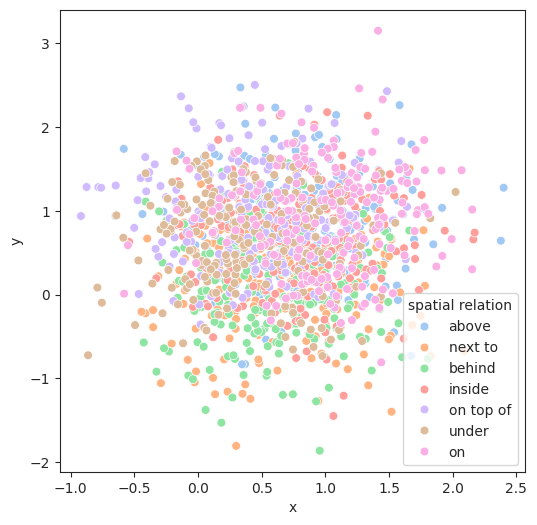}
	\end{minipage}
 \hspace{0.44cm}
	\begin{minipage}[t]{0.4\linewidth}
		\centering
		\includegraphics[scale = 0.26]{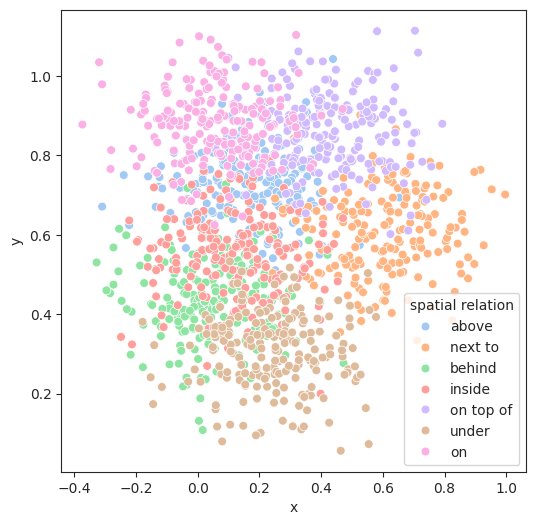}
	\end{minipage}
	\caption{Distribution of selected 200 samples across 7 common spatial relations with llava1.5 13B (acc 51.2\%) on the left and VSRE(acc 79.5\%) on the right.}
	\label{fig:scatter}
\end{figure}

To verify that the model has become more sensitive to positional information alongside the accuracy increase, we selected 7 common positional relationships. 
For each one, we sampled 200 instances and used sklearn's t-SNE for dimensionality reduction on visual tokens to plot the distribution of each relationship in Figure \ref{fig:scatter}.
It is evident that after dimensionality reduction, the visual features extracted by VSRE are better distinguished, with more pronounced inter-class differences. 
Notably, in the right scatter figure, semantically similar concepts such as ``on'' (pink), ``on top of'' (purple), and ``above'' (blue) are clustered together, while the contrasting ``under'' (brown) is far apart.

Furthermore, Table \ref{table:distance} presents the normalized average intra-class distance for each category. We use this metric to represent the model's ability to summarize and generalize single positional concepts; a smaller value indicates that the extracted features are better clustered (with more details  shown in Appendix Section5). 
It is evident that the visual features extracted by the more professional spatial sensor VSRE have a smaller average intra-class distance. 
This indicates that VSRE has a better understanding and generalization capability for various visual spatial concepts.

\begin{table}[]
\begin{tabular}{c|cccc}
\hline
Relations & above                & next to              & behind               & \multicolumn{1}{c}{inside} \\ \hline
LLaVA(51.2\%)     &0.54        &0.66                      & 0.53                     & 0.42       \\
VSRE(79.5\%)      &0.24        &0.34                      & 0.27                     &0.21        \\ \hline
Relations & on top of            & under                & on                   &AVG                            \\ \hline
LLaVA(51.2\%)     & 0.62 & 0.57   &0.69  &0.57                            \\
VSRE(79.5\%)      & 0.33 &0.29  &0.36  &0.29                            \\ \hline
\end{tabular}
\caption{Statistic result of average intra-class distance for each spatial relation category on 200 samples by llava1.5 13B (acc 51.2\%)  and VSRE(acc 79.5\%).}
	\label{table:distance}
\end{table}

\subsection{Bias Result}

\begin{figure}[htbp]
    \centering
    \begin{subfigure}
        \centering
        \includegraphics[scale = 0.15]{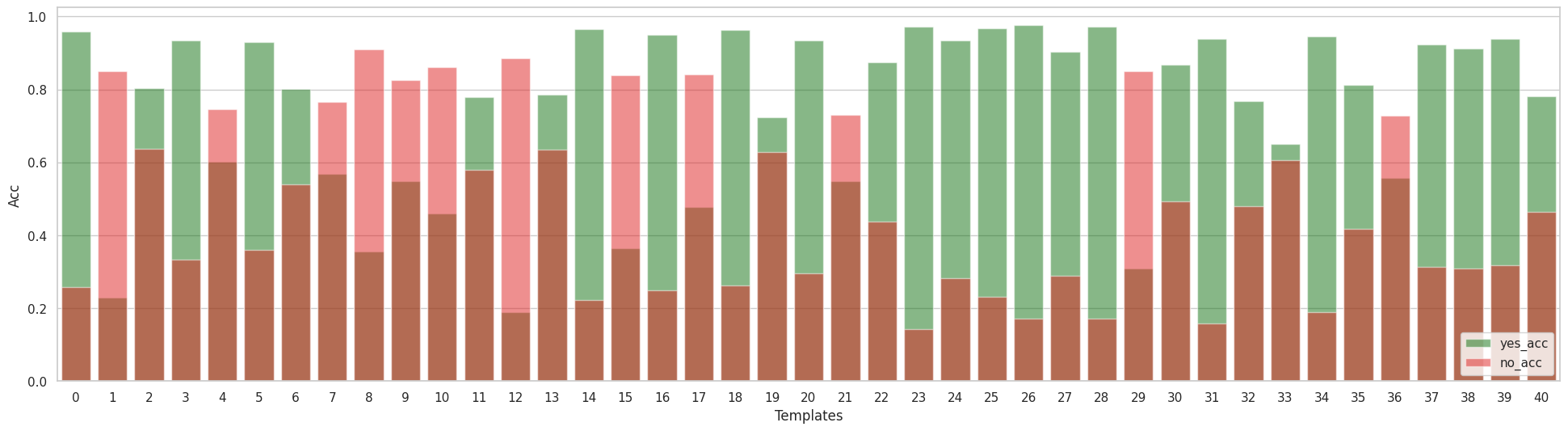}
    \end{subfigure}
    \par\bigskip 
    \begin{subfigure}
        \centering
        \includegraphics[scale = 0.15]{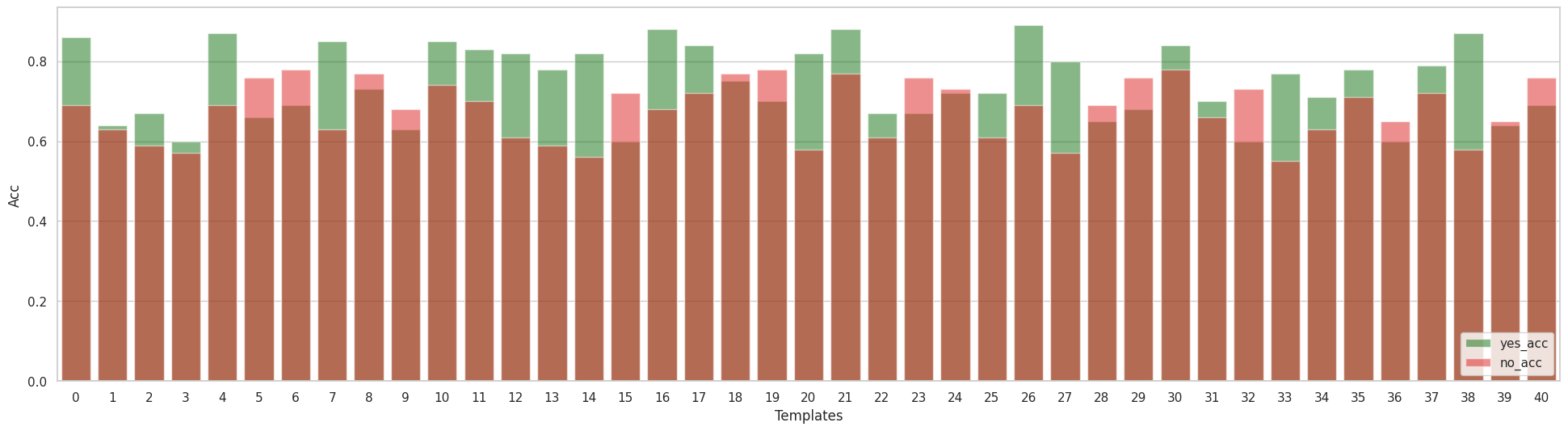}

    \end{subfigure}
    \caption{Comparison result between llava1.5 13B (acc 51.2\%) on the top and VSRE(acc 79.5\%) on the bottom with accuracy of yes question in green and no question in red. }
    \label{fig：yes_no}
\end{figure}

We re-testing VSRE ``yes'' and ``no'' question accuracy with the same template dataset in re-evaluation, and found that the response bias issue was alleviated in Figure \ref{fig：yes_no}.
Although the overall accuracy for ``yes'' questions is still higher than for ``no'' , the gap between the two has significantly narrowed.
As the accuracy of the red ``no'' questions has significantly improved, VSRE is not misled by the co-occurrence of related entity concepts in text and images but is more focused on the positional relationships between entities.

\section{Conclusion}
In this work, we first re-evaluate the VSR ability of VLLMs from scratch and diagnosed issues of inconsistent performance, hypersensitiveness on text prompts, lack of perception on visual details and answer bias. 
To address the problems, we first proposed the unified evaluation test set (\textit{Test-G} and \textit{Test-S}) for VLLM in QA scenario.
Next, we proposed methods for expanding both the data and the model structure. 
We validated the effectiveness of the methods through experiments on scaling data and scaling models.
We also validated the generalizability of our method on different benchmarks(MME, MMBench and SEED) and various models(VLLMs and LLMs).
Then, our proposed spatial expert VSRE surpasses the performance of LLaVA1.5 13B by 27\% (from 52\% to 79\%) on VSR test set.

Furthermore, in the sensitivity analysis, we found that while VSRE exhibited better differentiation of visual positional concepts, it also demonstrated a superior ability to summarize them.
A more sensitive visual position extractor brought about more specialized visual reasoning capabilities. This led the model to focus more on visual positional information, making it less influenced by the co-occurrence of entities and alleviating the issue of biased answer. Ultimately, our expansion method highlights the immense potential in VSR field for related data and model structures.

\section{Acknowledgements}
This work was supported by the National Key R\&D Program of China under grant 2023YFC3804600 and the Shenzhen Science and Technology Plan - Major Science and Technology Projects (No.KJZD20230923114213027)

\bibliography{aaai25}

\end{document}